\pdfoutput=1

\documentclass[11pt]{article}

\usepackage[final]{acl}

\usepackage{times}
\usepackage{latexsym}

\usepackage[T1]{fontenc}

\usepackage[utf8]{inputenc}

\usepackage{microtype}

\usepackage{inconsolata}

\usepackage{graphicx}

\usepackage{amsmath,amsfonts,bm}

\def\eqref#1{equation~\ref{#1}}

\def\1{\bm{1}}

\def\rmH{{\mathbf{H}}}

\def\rmQ{{\mathbf{Q}}}

\def\rmX{{\mathbf{X}}}

\def\rmZ{{\mathbf{Z}}}

\DeclareMathAlphabet{\mathsfit}{\encodingdefault}{\sfdefault}{m}{sl}
\SetMathAlphabet{\mathsfit}{bold}{\encodingdefault}{\sfdefault}{bx}{n}

\def\gH{{\mathcal{H}}}

\usepackage{url}
\usepackage{booktabs}
\usepackage{utfsym}
\usepackage{amsmath}
\usepackage{bbding}
\usepackage{amsfonts}
\usepackage{amssymb}
\usepackage{graphicx}
\usepackage{textcomp}
\usepackage{xcolor}
\usepackage{wrapfig}
\usepackage{color}
\usepackage{multirow}
\usepackage{pifont}
\usepackage{colortbl} 
\usepackage{tikz}
\usepackage{xcolor}
\usepackage{footnote}
\usepackage{tablefootnote}
\usepackage{graphicx}
\usepackage{wrapfig}
\usepackage{xcolor}
\definecolor{mydarkgreen}{RGB}{0,100,0} 

\usepackage{capt-of,etoolbox}

\title{Reminding Multimodal Large Language Models of Object-aware Knowledge with Retrieved Tags}

\author{
 \textbf{Daiqing Qi\textsuperscript{1}} \hspace{20pt}
 \textbf{Handong Zhao\textsuperscript{2}} \hspace{20pt}
 \textbf{Zijun Wei\textsuperscript{3}} \hspace{20pt}
 \textbf{Sheng Li\textsuperscript{1}}
\\
 \textsuperscript{1}University of Virginia \hspace{20pt}
 \textsuperscript{2}Adobe Research \hspace{20pt}
 \textsuperscript{3}Adobe Inc. 
\\
 \small{
   \texttt{\{daiqing.qi, shengli\}@virginia.edu} \hspace{10pt}
   \texttt{\{hazhao, zwei\}@adobe.com}
 }
}

\begin{document}

\makeatletter
\let\@oldmaketitle\@maketitle
\renewcommand{\@maketitle}{\@oldmaketitle
  \centering
  \includegraphics[width=0.93\textwidth,height=0.4\textheight,keepaspectratio]{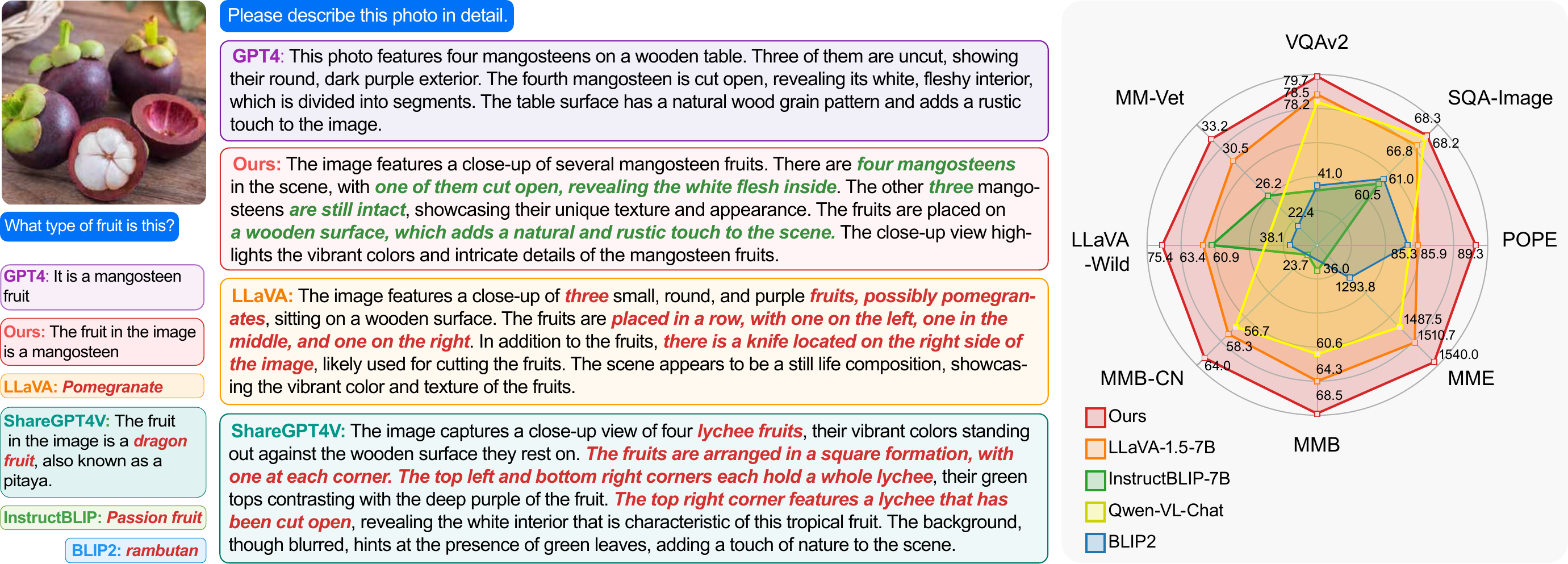}\bigskip
  \vspace{-5mm}
  \captionsetup{width=0.93\textwidth} 
  \captionof{figure}{Examples on LLaVA-W (\textbf{left}), and quantitative comparison (\textbf{right}). Imprecise low-quality answers are marked in \textcolor{red}{\textbf{\textit{red}}} and high-quality parts are marked in \textcolor{mydarkgreen}{\textbf{\textit{green}}}. 
  Popular \textit{open-source} MLLMs fail to identify the mangosteen (the first question), and list non-existent objects such as `knife' and incorrect quantities and arrangements, while ours correctly identify `mangosteens' with descriptions in detail.
  }
  \label{fig_example_1}
  \bigskip
  }
\makeatother

\maketitle

\begin{abstract}
Despite recent advances in general visual instruction-following ability of Multimodal Large Language Models (MLLMs), when diving into low-level details, they still struggle with critical problems when required to provide a precise and detailed response to a visual instruction:
(1) failure to identify novel \textit{objects} or entities, (2) mention of \textit{non-existent objects} and (3) neglect of \textit{object}'s attributed details.
Intuitive solutions include improving the size and quality of data or using larger foundation models. 
They show effectiveness in mitigating these issues, but at an expensive cost of collecting a vast amount of new data and introducing a significantly larger model.
Standing in the intersection of them, we examine the three \textit{object}-oriented problems from the 
perspective of the image-to-text mapping process by the multimodal connector.
In this paper, we first identify the limitations of multimodal connectors stemming from insufficient training data. Driven by it, we propose to enhance the mapping with retrieval-augmented tag tokens, which contain rich \textit{object}-aware information such as object names and attributes.
With our \textbf{T}ag-grounded visual instruction t\textbf{UN}ing with retrieval \textbf{A}ugmentation, \textbf{TUNA} outperforms baselines that share same language model and training data on 12 benchmarks. 
Furthermore, we show the zero-shot capability of TUNA when provided with specific datastores.

\end{abstract}


\section{Introduction}
\label{sec:intro}




Multimodal Large Language Models (MLLM) have witnessed remarkable progress recently~\cite{chen2023sharegpt4v, liu2023improved, liu2024visual, bai2023qwen, chen2023minigpt, dai2023instructblip, ye2023mplug, zhu2023minigpt, zhang2023llama}, exhibiting superior ability in following vision-and-language instructions. 
Despite their effectiveness in providing general responses, their performance often degrade when required to give a detailed and accurate answer to the question associated with an image with novel objects, named entities or complex scenes with rich and subtle details. 

Specifically, they frequently encounter challenges (Fig.~\ref{fig_example_1}) in: 1. identifying novel objects and named entities, 2. preventing the generation of objects that do not align with the target images, and 3. delivering a comprehensive description that covers the details of the target images.
We uncover the some of the potential causes of above challenges starting from the commonly adopted two-branch structure and the two-stage training paradigm of MLLMs: the first-stage pre-training and second-stage supervised fine-tuning (SFT). 
Most existing MLLMs such as LLaVA~\cite{liu2024visual} comprise two modules: (1) a vision branch consisting of a vision encoder and a multimodal connector, and (2) a Large Language Model (LLM).
In the pre-training stage with large-scale image-text pairs, the multimodal connector often learns to translate the outputs of the vision encoder to text embeddings, followed by the SFT stage which enhances the multi-modal instruction-following capabilities with instruction-format data.

\begin{figure}[t]
\vspace{-1mm}
\centerline{\includegraphics[scale=0.32]{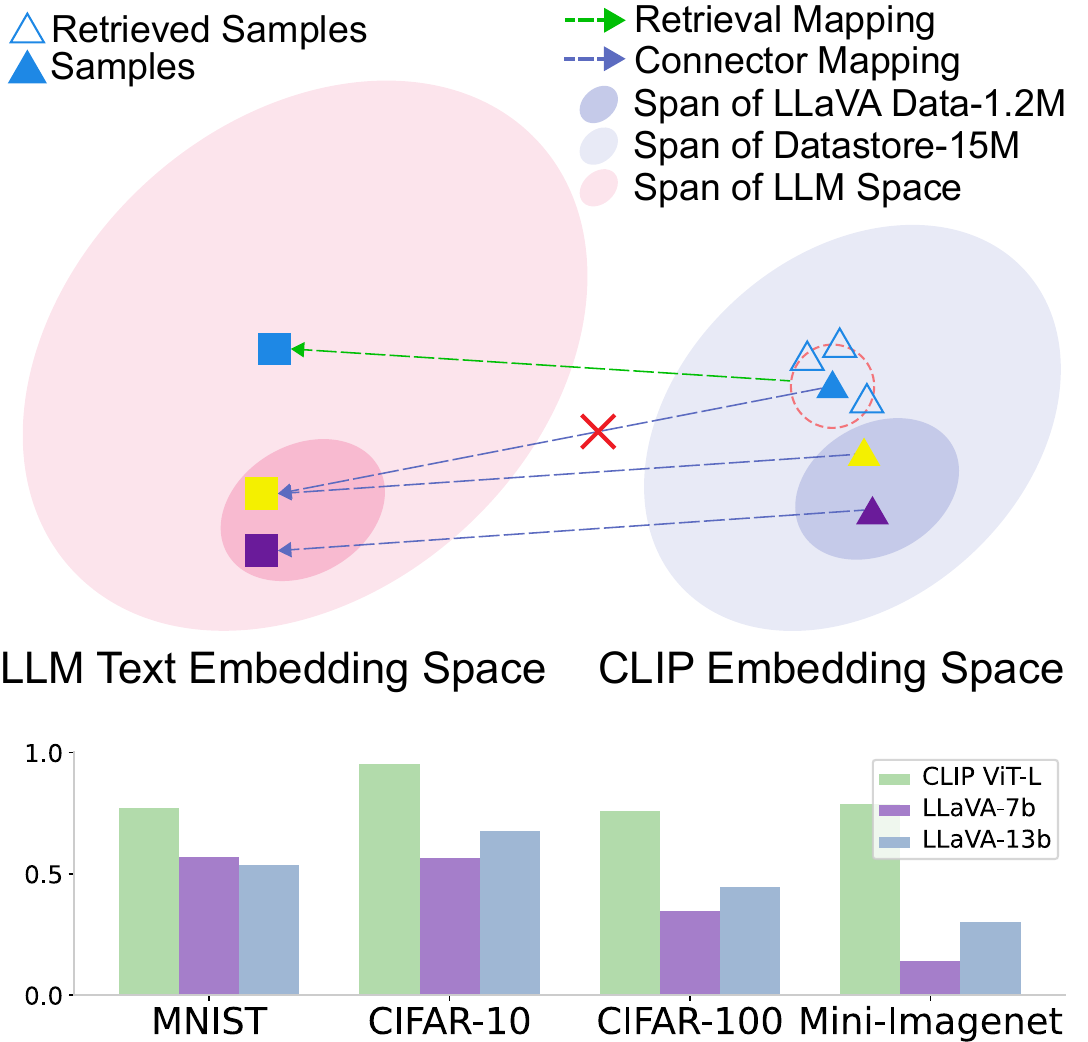}}
\caption{ \textbf{Top}: the process of translating image embeddings to text embeddings (LLaVA~\cite{liu2024visual}). \textbf{Bottom}: Image classification accuracy of CLIP~\cite{radford2021learning} and MLLMs built on it.} 
\label{fig_clip_plus}
\vspace{-7mm}
\end{figure}

Despite the promising zero-shot capability of the vision encoder, such as CLIP~\cite{radford2021learning}, which is pre-trained with over 400M image-text pairs, its generalizability is bottlenecked by the learnt mapping of the multimodal connector when integrated into the MLLM framework. 
E.g., in the case of LLaVA~\cite{liu2024visual}, the two-stage training data is significantly smaller compared to the pre-training data of its vision encoder CLIP (1.2M vs. 400M), as a result, the connector often fails to effectively map the out-of-distribution (OOD) images to the corresponding LLM text embeddings. Therefore, LLM fails to successfully identify image contents.
MLLMs' degradation on image classification performance~\cite{zhai2023investigating} is a simple illustration.
In Fig.~\ref{fig_clip_plus} (Bottom), an obvious classification performance gap between MLLMs and their frozen vision encoder (CLIP) is observed. 
The absence of similar classification objects in LLaVA's training data could be a critical factor, which makes it particular hard for the multimodal connector to translate OOD CLIP embeddings of test images to LLM text embeddings.

One intuitive solution is to enrich the training datasets with more image-text pairs, however, as high-quality instruction-format data is particularly critical for visual instruction tuning~\cite{chen2023sharegpt4v}, it is very expensive to build high-quality training data with hundreds of millions of image-text pairs of varying quality. Furthermore, the training could also become exceedingly burdensome.


Instead of directly improving the connector mapping with heavy training, could we build another lightweight new mapping as a complementary that effectively attends to objects, especially OOD ones?
Motivated by retrieval augmented generation (RAG)~\cite{ramos2023smallcap, ramos2023retrieval, yang2023re, hu2023reveal, lin2024fine, li2023evcap, yasunaga2022retrieval}, we propose a retrieval mapping.
As shown in Fig.~\ref{fig_clip_plus} (Top), while the connector fails to correctly map the sample out of LLaVA training data span to its corresponding text embedding in LLM embedding space (i.e., the blue triangle sample is incorrectly mapped to the yellow square sample)
, we introduce a large-scale external datastore with a better coverage of novel objects, named entities, and attributes, for the retrieval of useful knowledge towards the input image.
In this way, a new retrieval mapping could be built from the input image to corresponding LLM text embeddings (green dashed line in Fig.~\ref{fig_clip_plus}).

While most existing works retrieve relevant captions as extra knowledge, it may not apply here because all three challenges mentioned above are oriented with \textit{object}, where cleaner object-aware knowledge is urgent, instead of noisy captions. Therefore, we want to retrieve tags of the images that are similar to the input image as extra knowledge, where we can further enrich each tag representation with image region feature and adaptive weights to fulfill the potential of useful tags.
To this end, we introduce a \textbf{T}ag-grounded visual instruction t\textbf{UN}ing with retrieval \textbf{A}ugmentation, termed \textbf{TUNA}, that performs a knowledge-aware and tag-grounded generation. 
With grounded tags, TUNA is effective in identifying novel objects, named entities, and generate tag-oriented response which pays more attention to image details.

We summarize our contributions as follows: (\textit{i}) We identify potential factors hindering MLLMs and first propose a tag-grounded visual instruction tuning with retrieval-augmentation (TUNA) with enhanced knowledge on novel objects, more attention to details, and less mention of non-existent objects. (\textit{ii}) To fulfill the potential of tags, We carefully designed the image-aware tag encoder, which produces tag embeddings enhanced by image features with an adaptive weight. (\textit{iii}) We evaluate TUNA on extensive benchmarks along with a series of qualitative results, and show its zero-shot capability when provided
 with specific datastores.


\section{Related Works}
\noindent \textbf{Multimodal Large Language Models.} 
MLLMs evolve rapidly nowadays. With LLMs, while existing works~\cite{li2022blip, li2023blip} enable basic visual tasks like visual question answering, more recent works \cite{chen2023minigpt, liu2024visual} shows proficiency in image-text dialogues through alignment and fine-tuning. Subsequent research~\cite{bai2023qwen, chen2023shikra, dai2023instructblip, li2023otter, peng2023kosmos, ye2023mplug, you2023ferret} enhances LLMs by emphasizing data quality and diversity. With grounding data, a branch of works~\cite{ye2023mplug, you2023ferret, chen2023shikra, peng2023kosmos} improves LLMs’ grounding capability. 
Despite their evolution, as they share a similar multimodal connector module that performs image-to-text translation, a lingering fundamental problem persists: Out-of-distribution (OOD) images, such as novel objects, named entities, new scenes, etc., cannot be translated to text embeddings effectively, leading to misaligned answers, missing details or mention of non-existent objects from LLM.

\vspace{1mm}
\noindent \textbf{Retrieval-Augmented Multimodal Learning.} 
Retrieval-augmented language generation (RAG) consists of conditioning generation on additional information that is retrieved from an external datastore. Recently, A branch of works~\cite{ramos2023smallcap, ramos2023retrieval, yang2023re, hu2023reveal, lin2024fine, li2023evcap} integrate it into image captioning, where relevant captions are retrieved to guide the captioning.
Distinct from them, in visual instruction tuning, where detailed and dense responses based on the multimodal instructions are often required, cleaner object-level information, such as names and attributes of novel objects, named entities, is urgent. We provide a more detailed discussion in Appendix~\ref{sec:appendix:A}.

\noindent \textbf{Multimodal Learning with Tags.} Existing works~\cite{huang2023tag2text, zhou2020unified,li2020oscar,hu2021vivo,huang2022idea} show the effectiveness of introducing object tags as anchor points to help the learning of semantic alignments between images and texts in the training data. 
In the context of Fig.~\ref{fig_clip_plus}, they better align in-distribution data (yellow and purple samples) with tags. 
\textbf{Our goal is distinctive from them in that}, We do not aim to learn better representations of training data, instead, we want to (1) improve the tag-grounded generation capability of MLLMs and (2) acquire new knowledge with retrieved tags from \textit{external datastore}. 
Besides, as they treat object tags as anchor points for feature learning, tags are commonly human-used ones~\cite{huang2023tag2text} as guidance. For instance, Tag-to-Text~\cite{huang2023tag2text} collects 3,429 well-used tags filtered by human annotation. While in our case, where the large coverage is the priority, less frequently used tags (e.g., named entities) are also desired, resulting in a total of 3M tags. An expanded discussion is provided in Appendix~\ref{sec:appendix:A}.

\section{Tag-ground Visual Instruction Tuning}


In this section, we first introduce how we extract tags from 15M captions from CC12M~\cite{changpinyo2021conceptual} and CC3M~\cite{sharma2018conceptual}. Then we present how we build and use the datastore, followed by the illustration of TUNA.

\subsection{Multimodal Retriever}

\noindent \textbf{From Captions to Tags}.
As introduced in Sec.~\ref{sec:intro}, one of the fundamental challenges for MLLMs is to effectively translate image tokens to LLM text embeddings, especially for OOD images that contain novel objects.
With better translation, LLMs would be less likely to confuse with them, which could improve the identification of objects.
Thus in addition to the mapping learnt by the connector, we use a multimodal retriever to retrieve relevant information as an additional retrieval mapping (Fig.~\ref{fig_clip_plus}) to enhance the translation process.
Therefore, the quality of the retrieval mapping is critical. As a result, object-oriented tags as retrieved information would be very helpful. 
Additionally, with tag-grounded generation, retrieved tags also serve as groundings or hints, which could prompt the LLM to generate tag-aware contents \textit{if the tag is relevant to the input image}, which would also be helpful in alleviating missing objects or visual details.

Towards this end, we use CLIP image embeddings from image-text paired datasets as keys and corresponding \textit{tags} as values.
However, existing large-scale image-text datasets such as Conceptual Captions ~\cite{sharma2018conceptual, changpinyo2021conceptual} only contain captions. 
To mine tags from texts, we parse each caption into a set of tags with a combination of FACTUAL scene graph parser~\cite{li2023factual} and Name Entity Recognition (NER) with spaCy, yielding 3M tags extracted from 15M captions in CC3M~\cite{sharma2018conceptual} and CC12M~\cite{changpinyo2021conceptual}. We show several examples in Fig.~\ref{fig_parse}.
Details of the mining process are available in Appendix~\ref{sec:appendix:B}. 
We also provide a statistics of the obtained tags in Tab~\ref{tab_tag_sta}. 

\begin{figure}[t]
\vspace{-1mm}
\centerline{\includegraphics[scale=0.39]{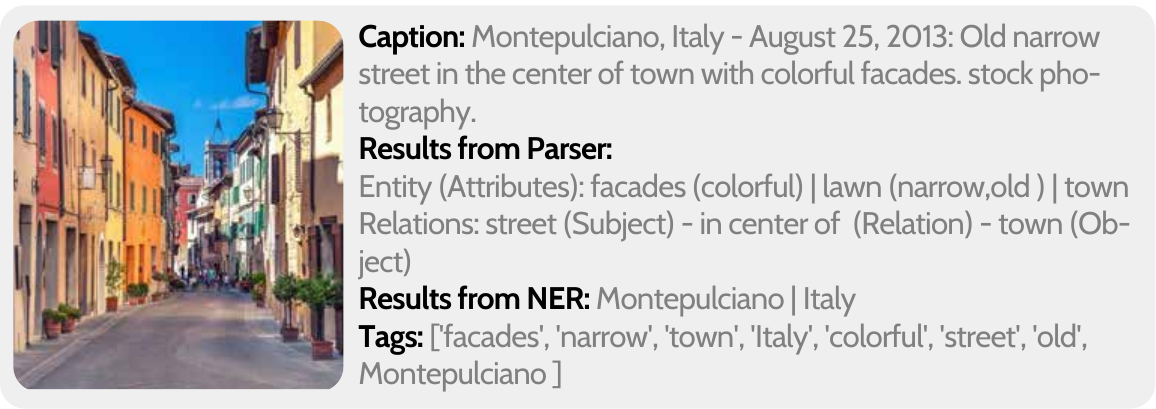}}
\caption{
Examples of tags derived from parsing and NER results.
}
\label{fig_parse}
\vspace{-4mm}
\end{figure}

\begin{table}[t]

\setlength{\tabcolsep}{5pt} 
\renewcommand{\arraystretch}{0.8} 

\centering
\scalebox{0.75}{
\begin{tabular}{cccccccc}
\toprule
 Number of Unique Tags & Characters per Tag & Tags per Image  & 
\\ 
\midrule
3.2M       &  16.8 & 5.31   \\
\bottomrule
\end{tabular}
}
\vspace{-3mm}
\caption{ Extracted tags from CC3M and CC12M}
\label{tab_tag_sta}
\vspace{-6mm}
\end{table}

\noindent \textbf{Datastore and Cross-Modal Retrieval}. 
With processed image-tags pairs, our datastore is indexed by FAISS library~\cite{johnson2019billion} with image CLIP embeddings as keys and associated tags as values. Given a query image, a \textit{k}-nearest neighbor retrieval with cosine similarity of embeddings between it and datastore images is performed. The \textit{tags} of top-\textit{k} retrieved images are input to TUNA as additional knowledge. 
In experiments, we use $k$=5.
We consider CC12M~\cite{changpinyo2021conceptual}, CC3M~\cite{sharma2018conceptual} and COCO~\cite{lin2014microsoft} training set as our datastore, resulting in 15M image-text pairs. In experiments, we use a whole combination, as well as parts of them, as our datastore to study how different datastores affect results. 
For Fashion QA, we use a combination of fashion data as our retrieval datastore.


\subsection{TUNA} 
\begin{figure*}[t]
\vspace{-1mm}
\centerline{\includegraphics[scale=0.33]{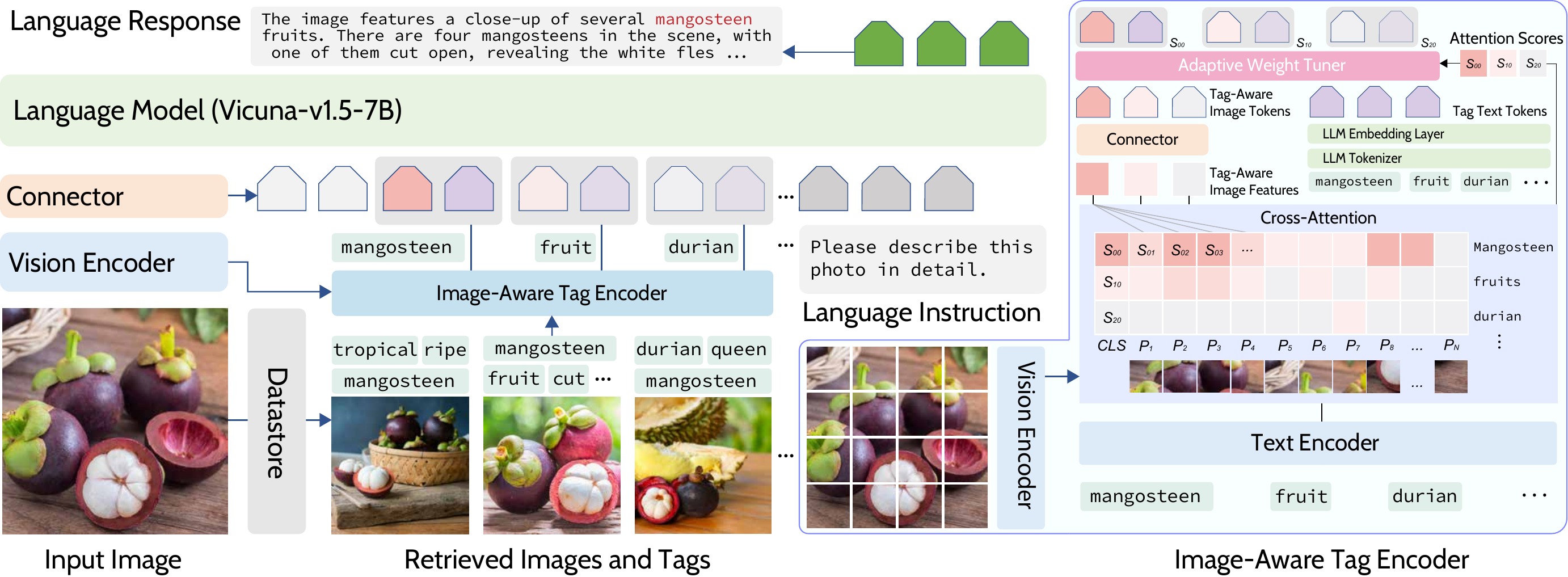}}
\caption{
 Framework of TUNA. 
\textbf{Left}: overall architecture. Given a language instruction, an image, and retrieved tags, they are transformed into tokens and input to the LLM. Only CLIP encoders are frozen.
\textbf{Right:} architecture of the image-aware tag encoder, which produces tag representations with retrieved tags and the input image.
}
\label{fig_framework}
\vspace{-4mm}
\end{figure*}

\noindent \textbf{Architecture}. 
The framework of TUNA is illustrated in Fig.~\ref{fig_framework}. Given a language instruction $\rmX_{\text{q}}$, and an input image $\rmX_{\text{v}}$, a set of images with associated tags are retrieved from the datastore. Assume there are $M$ tags in total, they are mixed together and denoted as $\{\rmX_{\text{t}}^{i}\}_{i=1}^{M}$. 
For image, a frozen pre-trained CLIP vision encoder ViT-L/14 is employed to extract the visual feature  $\rmZ_{\text{v}}=g(\rmX_{\text{v}}) \in \mathbb{R}^{[H \times W] \times D} $, followed by a MLP multimodal connector $h(\cdot)$ that translates the CLIP vision feature to text embeddings: $\rmH_{\text{v}}=h(\rmZ_{\text{v}})$. Similar to LLaVA~\cite{liu2024visual}, the grid visual features before the last Transformer layer are considered in our experiments.
The language instruction $\rmX_{\text{q}}$ is tokenized and projected to text embeddings $\rmH_{\text{q}}$ by the pre-trained LLM's tokenizer and embedding layer. 
Specifically, tags $\{\rmX_{\text{t}}^{i}\}^{M}_{i=1}$ are encoded by our image-aware tag encoder.

\vspace{2mm}
\noindent \textbf{Image-Aware Tag Encoder}. Given a tag $\rmX_{\text{t}}^{i}$, its tag representation $\gH_{\text{i}}$, which is encoded by our image-aware tag encoder, is a tuple of its text embedding $\rmH_{\text{t}}^{i}$ and the its tag-aware image token (embedding)  $\rmH_{\text{vt}}^{i}$, which contains visual features of the \textit{input query image} related to this tag. With this image token, LLM could better attend to details of the tag-related object in the input image.
Same with $\rmX_{\text{q}}$, the tag $\rmX_{\text{t}}^{i}$ is tokenized and projected to $\rmH_{\text{t}}^{i}$ with the LLM's tokenizer and embedding layer.
To obtain the tag-aware image token, the tag-aware image feature  $\rmZ_{\text{vt}}^{i} \in \mathbb{R}^{1 \times D}$ is first extracted from the grid visual features of the \textit{input image} via the cross-attention module: 
$
\rmZ_{\text{vt}}^{i}=\text{Cross-Att($\rmQ_{\text{t}}^{i}$, $\rmZ_{\text{v}}$, $\rmZ_{\text{v}}$)}
=\text{softmax}(\frac{\rmQ_{\text{t}}^{i} \rmZ_{\text{v}}^\intercal}{\sqrt{D}})\rmZ_{\text{v}}
$
,
where $\rmQ_{\text{t}}^{i} \in \mathbb{R}^{1 \times D}$ is the global CLIP text feature of tag $\rmX_{\text{t}}^{i}$, extracted by the frozen CLIP text encoder. 
Then we obtain the tag-aware image token $\rmH_{\text{vt}}^{i}=h(\rmZ_{\text{vt}}^{i})$. Finally, the tag representation $\gH_{i}$ consists of the tuple $(\rmH_{\text{vt}}^{i}, \rmH_{\text{t}}^{i})$.
Iterating over all tags, we have $\{\gH_{i}\}_{i=1}^{M}$.

\vspace{2mm}
\noindent \textbf{Adaptive Weight Tuner}.
As retrieved images may contain less relevant or irrelevant tags, e.g., the tag \texttt{durian} in Fig~\ref{fig_framework}, we apply an adaptive weight tuner over them to give more attention to highly relevant tags while ignoring less related ones. 
Specifically, the score of $\gH_{i}$ is the cosine similarity between $\rmQ_{\text{t}}^{i}$ and the global CLIP visual feature (i.e., the <CLS> token) of the input image. The scores are normalized to [0,1] as the final weights, which are applied to $\rmH_{\text{vt}}^{i}$ and $\rmH_{\text{t}}^{i}$ before input to the LLM.
Despite its simplicity, this approach proves to be effective in Tab.~\ref{tab_aba}.

\vspace{2mm}
\noindent \textbf{Supervised Fine-Tuning}. 
We consider Vicuna-7B~\cite{chiang2023vicuna}, a decoder-only LLM instruction-tuned on top of LLaMA~\cite{touvron2023llama}, as our language model. 
We use both image and text encoders from CLIP-ViT-L/14@336p.
We initialize the pre-trained multimodal connector from LLaVA-1.5~\cite{liu2023improved}. During the instruction tuning, we always keep the weights of the vision encoder frozen, and update both the pre-trained weights of the connector and the LLM.

\section{Experiment}
In this section, we first present the training details of TUNA and benchmarks. Then we introduce quantitative and qualitative comparison with popular open-source models, followed by detailed analysis experiments and ablation studies. 

\vspace{1mm}
\noindent \textbf{Training Details.} TUNA is finetuned on instruction data for one epoch, following existing works~\cite{liu2023improved,chen2023sharegpt4v}. 
We consider two different instruction-following datasets in our experiments: LLaVA-665K~\cite{liu2023improved} and ShareGPT4V-665K~\cite{chen2023sharegpt4v} as our instruction-following data during fine-tuning separately, resulting in two versions of our model, TUNA and $\text{TUNA}^{+}$. ShareGPT4V-665K contains instruction-following data with higher quality. Details on datasets are available in Appendix~\ref{sec:appendix:C}.
We apply a learning rate of 2$e$-5 and a batch size of 128. The training takes 12$\sim$14 hours with 8 A100 GPUs with ZeRO3. Details are available in Appendix~\ref{sec:appendix:C}. 

\vspace{1mm}
\noindent \textbf{Benchmarks.} 
We compare TUNA with baselines on 12 benchmarks, including VQA benchmarks and multimodal benchmarks designed for LLMs. Details are available in Appendix~\ref{sec:appendix:G}.
\begin{table*}[t]

\setlength{\tabcolsep}{3pt} 
\renewcommand{\arraystretch}{0.8} 

\begin{center}
\scalebox{0.65}{
\begin{tabular}{llll|lllll|lllllll}
\toprule
Method & LLM & V-Enc. & IT & $\mathrm{VQA}^{\mathrm{v} 2}$ & GQA & VizWiz & $\mathrm{SQA}^{\mathrm{I}}$ & $\mathrm{VQA}^{\mathrm{T}}$ & POPE & MME & $\mathrm{MMB}$ & $\mathrm{MMB}^{\mathrm{CN}}$ & SEED & $\mathrm{LLaVA}^{\mathrm{W}}$ & MM-Vet \\
\midrule
BLIP-2 & Vicuna-13B & - & - & 41.0 & 41.0 & 19.6 & 61.0 & 42.5 & 85.3 & 1293.8 & - & - & 46.4 & 38.1 & 22.4 \\
InstructBLIP & Vicuna-7B & - & 1.2M & - & 49.2 & 34.5 & 60.5 & 50.1 & - & - & 36 & 23.7 & 53.4 & 60.9 & 26.2 \\
InstructBLIP & Vicuna-13B & - & 1.2M & - & 49.5 & 33.4 & 63.1 & 50.7 & 78.9 & 1212.8 & - & - & - & 58.2 & 25.6 \\
Shikra & Vicuna-13B & - & 5.5M & 77.4 & - & - & - & - & - & - & 58.8 & - & - & - & - \\
IDEFICS-9B & LLaMA-7B & - & 1M & 50.9 & 38.4 & 35.5 & - & 25.9 & - & - & 48.2 & 25.2 & - & - & - \\
IDEFICS-80B & LLaMA-65B & - & 1M & 60.0 & 45.2 & 36.0 & - & 30.9 & - & - & 54.5 & 38.1 & - & - & - \\
Qwen-VL & Qwen-7B & - & 50M & 78.8 & 59.3 & 35.2 & 67.1 & 63.8 & - & - & 38.2 & 7.4 & 56.3 & - & - \\
Qwen-VL-Chat & Qwen-7B & - & 50M & 78.2 & 57.5 & 38.9 & 68.2 & 61.5 & - & 1487.5 & 60.6 & 56.7 & 58.2 & - & - \\
\textcolor{gray}{ShareGPT4V} & \textcolor{gray}{Vicuna-13B} & $\text{CLIP}^{\text{V-L}}_{336}$ & \textcolor{gray}{665K(S)} & \textcolor{gray}{81.0} & \textcolor{gray}{63.4} & \textcolor{gray}{55.6} & \textcolor{gray}{71.2} & \textcolor{gray}{62.2} & \textcolor{gray}{85.9} & \textcolor{gray}{1618.7} & \textcolor{gray}{68.5} & \textcolor{gray}{63.7} & \textcolor{gray}{70.8} & \textcolor{gray}{79.9} & \textcolor{gray}{43.1} \\
\textcolor{gray}{LLaVA-1.5} & \textcolor{gray}{Vicuna-13B} & $\text{CLIP}^{\text{V-L}}_{336}$ & \textcolor{gray}{665K(L)} & \textcolor{gray}{80.0} & \textcolor{gray}{63.3} & \textcolor{gray}{53.6} & \textcolor{gray}{71.6} & \textcolor{gray}{61.3} & \textcolor{gray}{85.9} & \textcolor{gray}{1531.3} & \textcolor{gray}{67.7} & \textcolor{gray}{63.6} & \textcolor{gray}{61.6} & \textcolor{gray}{70.7} & \textcolor{gray}{35.4} \\
\textcolor{gray}{LLaVA-1.6/NeXT} & \textcolor{gray}{Vicuna-7B} & $\text{CLIP}^{\text{V-L}}_{ \text{\textcolor{blue}{\textbf{AR}}}}$ &\textcolor{gray}{760K(N)} & \textcolor{gray}{81.8} & \textcolor{gray}{64.2} & \textcolor{gray}{57.6} & \textcolor{gray}{70.1} & \textcolor{gray}{64.9} & \textcolor{gray}{86.5} & \textcolor{gray}{1519.0} & \textcolor{gray}{67.4} & \textcolor{gray}{60.6} & \textcolor{gray}{70.2} & \textcolor{gray}{81.6} & \textcolor{gray}{43.9} \\
\midrule
ShareGPT4V & Vicuna-7B & $\text{CLIP}^{\text{V-L}}_{336}$ & 665K(S) & 80.6 & \text{63.3} & 57.2 & 68.4 & \text{60.4} & 85.3 & 1567.4 & 68.8 & 62.2 & 69.7 & 72.6 & 37.6 \\
\rowcolor{gray!20} $\text{Ours}^{+}$ & Vicuna-7B & $\text{CLIP}^{\text{V-L}}_{336}$ & 665K(S) & \textbf{81.1} & \textbf{63.4} &\textbf{57.4}& \textbf{70.8} &\textbf{60.4}& \textbf{89.6} & $\textbf{1583.8}$ & \textbf{70.8} & \textbf{65.0} & $\textbf{70.6}$ & $\textbf{80.1}$ & \textbf{40.1} \\
\midrule
LLaVA-1.5 & Vicuna-7B & $\text{CLIP}^{\text{V-L}}_{336}$ & 665K(L) & 78.5 & 62.0 & 50.0 & 66.8 & 58.2 & 85.9 & 1510.7 & 64.3 & 58.3 & 58.6 & 63.4 & 30.5 \\
\rowcolor{gray!20}Ours & Vicuna-7B & $\text{CLIP}^{\text{V-L}}_{336}$ & 665K(L) & \textbf{79.7} & \textbf{62.6} &\textbf{50.0}& \textbf{68.3} &\textbf{58.4}& \textbf{89.5} & $\textbf{1540.0}$ & \textbf{68.5} & \textbf{64.0} & $\textbf{59.6}$ & $\textbf{75.4}$ & \textbf{33.2} \\
\bottomrule
\end{tabular}
}
\end{center}
\caption{
\textbf{Comparison with SoTA methods on 12 benchmarks.} Our model achieves the best performance on 12 benchmarks compared with LLMs that are finetuned from the same instruction 
tuning (IT) datasets with the same configuration on the vision encoder (V-Enc.) and language model (Vicuna-7B). Best results are in \textbf{bold}.
} 
\label{tab_main}
\end{table*}

\subsection{Comparison with Baselines}
\noindent \textbf{Main Results.} 
In Tab.~\ref{tab_main}, we provide a quantitative comparison of TUNA with popular open-source MLLMs.
On 12 benchmarks, TUNA consistently outperforms previous LLMs that are finetuned from the same instruction-tuning datasets as ours with the same configuration on the vision encoder and language model (Vicuna-7B), especially on recent multimodal benchmarks with more notable improvements.
As the size of LLM and different choices of instruction-following data can significantly improve the model performance,
we mark the models gray that are equipped with a larger 13B language model or finetuned from currently unavailable datasets of higher quality and quantity.
Specifically, LLaVA-1.6 (or LLaVA-NeXT)\footnote{https://llava-vl.github.io} is finetuned from larger instruction-following data of higher quality, with additional user instruct data. Besides, it equips the better vision encoder with dynamic high resolution, known as AnyRes (AR).
Although it is not a fair comparison, we still outperform LLaVA-1.6 in $\text{MMB}^{\text{CN}}$, MMB and POPE, and the corresponding 13B models in $\text{MMB}^{\text{CN}}$, MMB, POPE and LLaVA-W.

\begin{figure}[t]
\vspace{-1mm}
\centerline{\includegraphics[scale=0.28]{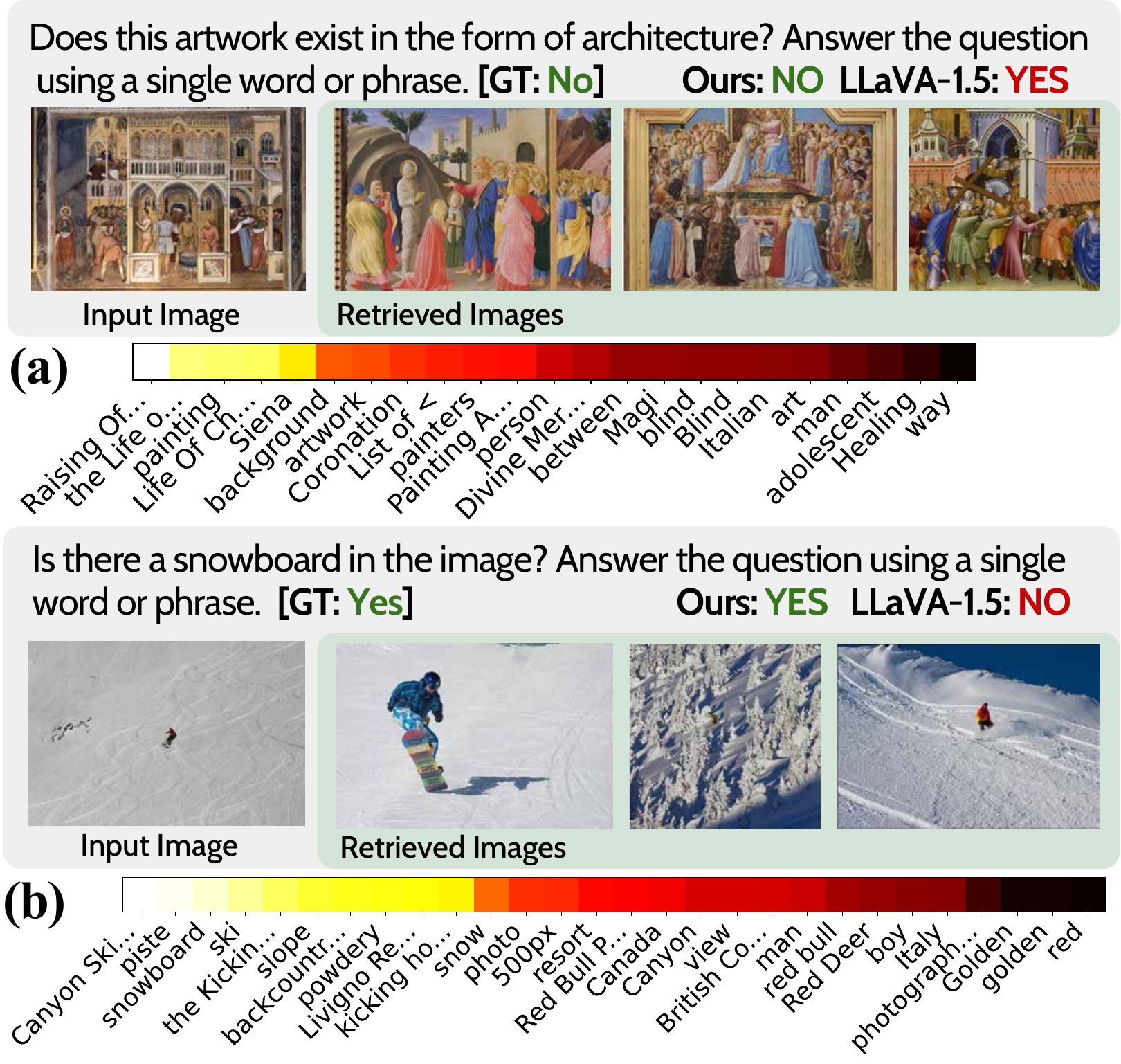}}
\caption{
\textbf{VQA examples of TUNA.} For each example, we show top 3 retrieved images to save space. We show all tag set associated with all retrieved images as well as their tuned weights in heat map, where the brightest region for the highest weight 1 and darkest region for the lowest weight 0 (Zoom in for better view). Correct answers are marked green and wrong ones in red. More examples are available in Appendix~\ref{sec:appendix:E}.
}
\label{fig_QACase}
\vspace{-6mm}
\end{figure}

\noindent \textbf{How Can TUNA Improve the Recognition of Novel Objects and Entities?} As visualized in Fig.~\ref{fig_clip_plus} (Top), with our 15M large-scale datastore, the new retrieval mapping could greatly compensate for the original LLaVA multimodal connector that learns from around 1M data. With the additional mappings from retrieval data, TUNA is expected to show particularly improvements over questions towards novel objects or entities in the given input image. 
We show sub-tasks from MME~\cite{fu2023mme} and MMB~\cite{liu2023mmbench} that consists of such questions in Tab.~\ref{tab_identi}. We gain obvious improvements over the baseline in most sub-tasks. 
We also show several VQA examples from multimodal benchmarks in Fig~\ref{fig_example_1} and Fig~\ref{fig_QACase}.
In Fig~\ref{fig_example_1}, all of the baselines fail to correctly identify this fruit as a mangosteen, including LLaVA-1.5. It is reasonable as mangosteens do not appear in the its training data, which makes it particularly hard for the connector to map it to somewhere close to text embeddings of ``mangosteen'' in the LLM embedding space, as illustrated in Fig~\ref{fig_clip_plus}. 
When the question about the given image is a little tricky, e.g., in Fig~\ref{fig_QACase} (a), the MLLM is asked if a painting of a building exists in the form of architecture, LLaVA-1.5 is confused on whether it is a real architecture or a painting. However, TUNA easily distinguished it from real architectures with additional knowledge from retrieved tags of similar images in datastore.

\begin{table}[t]

\setlength{\tabcolsep}{3.5pt} 
\renewcommand{\arraystretch}{0.8} 

\centering
\scalebox{0.65}{
\begin{tabular}{llccccccc}
\toprule
 Model & Posters & Celebrity & Artwork & landmark & Image Style & Celeb &
\\ 
\midrule
LLaVA-1.5       & 146.6   & 137.1     & 119.5  & 163.8    & 69.1    & 83.8  \\
Ours    & \textbf{155.9}   & \textbf{154.7}     & \textbf{128.7}  & \textbf{166.3}  &\textbf{81.1}& \textbf{85.8}  \\
\bottomrule
\end{tabular}
}
\caption{Results on sub-tasks of MME~\cite{fu2023mme} and MMB~\cite{liu2023mmbench}, where questions are towards novel objects, entities or scenes in the image. Otherwise mentioned, backbone LLM is Vicuna-7B.} 
\label{tab_identi}
\end{table}

\begin{table}[t]

\setlength{\tabcolsep}{2pt} 
\renewcommand{\arraystretch}{0.8} 

\begin{center}
\scalebox{0.65}{%

\begin{tabular}{l|l|ccccc}
\toprule
Datasets & Metrics & \cellcolor{gray!15}Ours & Ferret & InstructBLIP & LLaVA & mPLUG-Owl \\
\midrule 
\multirow{4}{*}{ Random } & Accuracy $(\uparrow)$ & \cellcolor{gray!15}\textbf{91.00} & 90.24 & 88.57 & 88.00 & 53.97 \\
& Precision $(\uparrow)$ & \cellcolor{gray!15}\textbf{98.05} & 97.72 & 84.09 & 97.44 & 52.07 \\
& Recall $(\uparrow)$ & \cellcolor{gray!15}\textbf{84.10} & 83.00 & 95.13 & 78.80 & 99.60 \\
& F1 Score $(\uparrow)$ & \cellcolor{gray!15}\textbf{90.93} & 89.76 & 89.27 & 87.13 & 68.39 \\
\midrule 
\multirow{4}{*}{ Popular } & Accuracy $(\uparrow)$ & \cellcolor{gray!15}\textbf{90.16} & 84.90 & 82.77 & 87.43 & 50.90 \\
& Precision $(\uparrow)$ & \cellcolor{gray!15}\textbf{95.46} & 88.24 & 76.27 & 95.24 & 50.46 \\
& Recall $(\uparrow)$ & \cellcolor{gray!15}\textbf{84.20} & 80.53 & 95.13 & 78.80 & 99.40 \\
& F1 Score $(\uparrow)$ & \cellcolor{gray!15}\textbf{90.56} & 84.21 & 84.66 & 86.24 & 66.94 \\
\midrule 
\multirow{4}{*}{ Adversarial } & Accuracy $(\uparrow)$ & \cellcolor{gray!15}\textbf{88.43} & 82.36 & 72.10 & 85.50 & 50.67 \\
& Precision $(\uparrow)$ & \cellcolor{gray!15}\textbf{91.99} & 83.60 & 65.13 & 90.99 & 50.34 \\
& Recall $(\uparrow)$ & \cellcolor{gray!15}\textbf{84.20} & 80.53 & 95.13 & 78.80 & 99.33 \\
& F1 Score $(\uparrow)$ & \cellcolor{gray!15}\textbf{87.63} & 82.00 & 77.32 & 84.45 & 66.82 \\
\midrule 
\multicolumn{2}{c|}{Average F1} & \cellcolor{gray!15}\textbf{89.50} & 85.32 & 83.75 & 85.94 & 67.38 \\
\bottomrule
\end{tabular}
}
\end{center}
\caption{ Results on POPE. 
We show most competing baselines. Full table is available in Appendix~\ref{sec:appendix:F}.
TUNA outperform Ferret~\cite{you2023ferret}, which is finetuned on grounding and referring data. }
\vspace{-6mm}
\label{tab_POPE}
\end{table}

\begin{figure}[t]
\centerline{\includegraphics[scale=0.35]{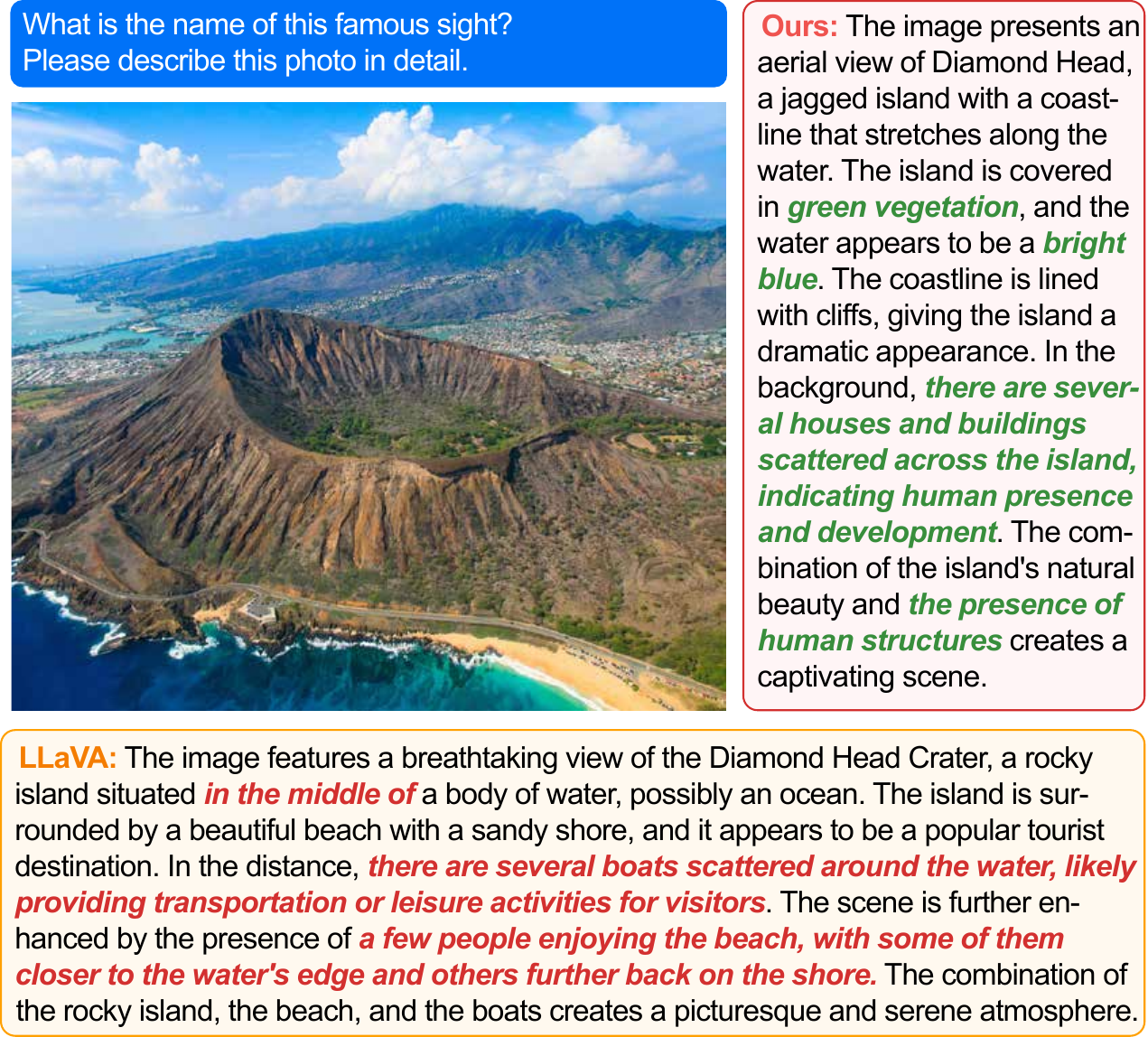}}
\caption{
\textbf{TUNA on LLaVA-W examples}. Imprecise low-quality answers are marked in \textcolor{red}{\textbf{\textit{red}}} and high-quality parts are marked in \textcolor{mydarkgreen}{\textbf{\textit{green}}}. TUNA does not mention non-existent objects and gives a more detailed description.
}
\label{fig_detailed_example}
\vspace{-6mm}
\end{figure}

\vspace{1mm}
\noindent \textbf{How Can TUNA Help to Identify the Existence of Objects?}
With an input image, the retrieved images are often similar to it or in the similar context. Intuitively, the retrieved images are very likely to contain similar elements or objects to the input image. Therefore, the tags could be helpful to provide additional hints to the LLM to pay special attention to them about their existence.
We evaluate our model on POPE~\cite{li2023evaluating}, a benchmark designed towards the existence of objects.
Results are available in Tab.~\ref{tab_POPE}, we outperform competing baselines including referring and grounding MLLMs such as Ferret~\cite{you2023ferret} and Shikra~\cite{chen2023shikra}.
A simple example is shown in Fig~\ref{fig_QACase} (b). When the object is visually imperceptible in the image, hints from tags are very helpful.

\vspace{1mm}
\noindent \textbf{How Can TUNA Attend to Rich Details with Tags?} While most of the traditional VQA benchmarks and multimodal benchmarks provide short questions answering pairs (Fig~\ref{fig_QACase}), LLaVA-W~\cite{liu2023improved} evaluates MLLM's capability of giving long detailed response. 
Quantitative results are available in Tab.~\ref{tab_llava_w}. TUNA consistently outperforms baselines. We also provide one example in Fig.~\ref{fig_detailed_example}.
While LLaVA mentions non-existent boats, people, TUNA accurately describes the water body, the existence of green vegetation, and interestingly, the presence of houses and buildings behind the mountain (zoom in for better view). More interestingly, there are no retrieved noun tags directly related ``houses'' or ``buildings''. By removing tags one by one, we finally identify that the tag ``accessible'' contributes to the the description of houses and buildings. 
It is an interesting phenomenon that somehow tells us that not only nouns can remind the LLM the existence of objects, relevant adjectives can also teach the LLM to pay attention to visual details. In this case, ``accessible'' means ``human can access to this place'', which might remind the LLM the existence of houses and buildings.


\subsection{Ablation Study}
\vspace{1mm}
\noindent \textbf{Ablation of Adaptive Weight Tuner.} Grounded on tags, intuitively, the quality of tags is critical to TUNA. 
However, retrieved tags could be noisy.  E.g., the tag \texttt{durian} in Fig~\ref{fig_framework}.
To this end, we apply an adaptive weight tuner in our image-aware tag encoder to allocate more weight to more relevant tags and less weight to less relevant ones. 
We first ablate the tuner module to show its effectiveness of this simple but critical component in alleviating the noises of tags. Without the adaptive weight tuner, all retrieved tags would be equal important and their weights are set to the maximum value.
The result is shown in Tab.~\ref{tab_aba} (w/o tuner). 
A clear performance drop is observed compared to the full method.
It is reasonable because while related tags can provide useful information to the LLM, the irrelevant tags are misleading.
Although it underperforms the full method, without the tuner, our model is still comparable or slightly better than LLaVA-1.5. This is favourable because it manifests that our model itself is somehow robust against less relevant tags without the tuner.

\vspace{1mm}
\noindent \textbf{Effectiveness of Instruction Tuning.}
Since MLLMs are naturally in-context learners, we are interested in the effectiveness of our tag-grounded finetuning compared to the vanilla LLaVA-1.5, where tags are provided as in-context knowledge. 
For fair comparison, we apply the weight tuner to both models. Let's refer this model as $\text{TUNA}^{-}$.
Results in Fig.~\ref{tab_aba} (w/o FT) indicates that, the LLM without tag-grounded instruction tuning cannot make effective use of informative tags.

\begin{table}[t]

\setlength{\tabcolsep}{6pt} 
\renewcommand{\arraystretch}{0.8} 

\begin{center}
\scalebox{0.7}{%
\centering
\begin{tabular}{lccccc}
\toprule
Model  & Average & Conversation & Reasoning & Detail \\ 
\midrule
\textcolor{gray}{ShareGPT4V-13B} & \textcolor{gray}{79.5} & \textcolor{gray}{81.4} & \textcolor{gray}{79.2} & \textcolor{gray}{76.2} \\
\textcolor{gray}{LLaVA-v1.5-13B} & \textcolor{gray}{72.9} & \textcolor{gray}{82.8} & \textcolor{gray}{74.3} & \textcolor{gray}{53.1} \\ 
\midrule
ShareGPT4V-7B & 74.9 & 78.5 & 69.2 & 74.4 \\ 
\rowcolor{gray!15} $\text{Ours}^{+}$ & \textbf{80.1} & \textbf{87.0}& \textbf{80.2} & \textbf{77.2} \\ 
\midrule
LLaVA-v1.5-7B  & 65.3 & 81.3 & 64.0 & 52.9 \\ 
\rowcolor{gray!15} Ours & \textbf{75.4} & \textbf{82.0} & \textbf{77.2} & \textbf{62.5} \\ 
\bottomrule
\end{tabular}
}
\end{center}
\vspace{-2mm}
\caption{Results on LLaVA-in-the-Wild (LLaVA-W) Bench. Our model consistently outperforms baselines that share the same LLM and instruction tuning data.}
\label{tab_llava_w}
\vspace{-6mm}
\end{table}

\vspace{1mm}
\noindent \textbf{Are Tags more Effective than Sentences?}
We compare TUNA with sentence-level retrieval in Tab.~\ref{tab_aba} (w/ captions). Instead of tags, we finetune TUNA with captions of retrieved images as additional knowledge. The image-aware tag encoder is also used, but the input tags are replaced by captions. Results show that sentence-level retrieval is not helpful. It is reasonable because tags provide cleaner and more object-related knowledge such as names, attributes, while captions are noisy.

\vspace{1mm}
\noindent \textbf{Would Irrelevant Tags Hurt the Backbone during Inference?} 
It is intuitive that a large-scale datastore often covers useful knowledge to the input image and question. Therefore, useful tags could be retrieved. However, there might be corner cases when retrieved tags are all irrelevant. 
To this end, we run experiments without tags and with random tags. Results are reported in Tab.~\ref{tab_aba}. With irrelevant tags, TUNA is comparable to its backbone LLaVA-1.5.
It manifests that, our method notably improves the backbone performance with useful tags and will not hurt the backbone performance when only irreverent tags are available.

\begin{table}

\setlength{\tabcolsep}{6pt} 
\renewcommand{\arraystretch}{0.8} 

\begin{center}
\scalebox{0.65}{%
\centering
\begin{tabular}{l|ccccc}
\toprule
Method & POPE & $\mathrm{MMB}^{\mathrm{CN}}$ & MMB & MM-VET & LLaVA-W \\
\midrule
\rowcolor{gray!15}Full method  &\textbf{89.5} &\textbf{64.0} & \textbf{68.5} &\textbf{33.2} &\textbf{75.4} \\
\midrule
\hspace{0.1cm}  w/o tuner  & 86.9 & 58.2 & 64.2 & 31.8 & 65.7 \\
\hspace{0.1cm}  w/o tags  & 85.9 & 58.6 & 64.9 & 31.2 & 65.6 \\
\hspace{0.1cm}  w/ random tags & 85.3 & 58.0 & 64.7 & 31.1 & 65.0 \\
\midrule
\hspace{0.1cm}  w/o FT ($\text{TUNA}^{-}$) & 85.9 & 58.1 & 64.2 & 30.8 & 66.5 \\
\midrule
\hspace{0.1cm}  w/ captions & 85.5 & 59.3 & 65.4 & 30.6 & 65.7 \\
\midrule
LLaVA-1.5 & 85.3 & 58.2 & 64.3 & 30.5 & 65.3 \\
\bottomrule
\end{tabular}
}
\end{center}
\vspace{-3mm}
\caption{Ablation Studies on (1) the effectiveness of the adaptive weight tuner, (2) retrieved tags during inference and (3) tag-grounded finetuning.} 
\label{tab_aba}
\vspace{-3mm}
\end{table}

\vspace{1mm}
\noindent \textbf{Different Choices of Datastore.}
We also study how different choices of datastores can affect the model performance. In the default setting, we use a combination of CC12M, CC3M and COCO training set. 
In addition, we perform the tag-grounded instruction tuning with different datastores, and use them for retrieval during inference, respectively. Results are available in Tab.~\ref{tab_aba_datastore}. Default setting with largest datastore size outperforms other baselines. We have detailed analysis in Appendix~\ref{sec:appendix:H}.
\vspace{-2mm}

\begin{table}

\setlength{\tabcolsep}{5pt} 
\renewcommand{\arraystretch}{0.8} 

\begin{center}
\scalebox{0.7}{%
\centering
\begin{tabular}{l|ccccc}
\toprule
Datastore & POPE & MMB & $\mathrm{MMB}^{\mathrm{CN}}$  & MM-VET & LLaVA-W \\
\midrule
\rowcolor{gray!15} All   &\textbf{89.5} &\textbf{68.5} &\textbf{64.0} &\textbf{33.2} &\textbf{75.4} \\
\midrule
 CC12M                                & 86.6 & \underline{67.8} & \underline{63.4} &  \underline{32.6} & \underline{73.8} \\
 CC3M                                 & 86.2 & 67.5 & 62.9 & 32.1 & 69.2 \\
 $\text{COCO}^{\text{}}$         & \underline{87.9} & 65.9 & 60.2 & 31.4 & 65.2 \\
 \midrule
  w/o Datastore                           & 85.3 & 64.3  & 58.2 & 30.5 & 65.3 \\
\bottomrule
\end{tabular}
}
\end{center}
\vspace{-3mm}
\caption{Ablations on the choice of datastores.}
\label{tab_aba_datastore}
\vspace{-6mm}
\end{table}


\subsection{Zero-shot Inference on Fashion Domain}
\noindent \textbf{Fashion-Bench.}
To study TUNA on OOD data from another specific domain, we further collect data from FashionGen~\cite{rostamzadeh2018fashion} validation set and create a benchmark to measure the model’s instruction-following capability in fashion domain, similar to LLaVA-Bench.
Following LLaVA, we also collect a set of 24 images from FashionGen, with one question associated with each image.
The questions are from one of the three types:
(1) \textit{Conversation}.
    We design a conversation between the assistant and a person asking questions about the product, including the product brands, categories, materials, etc. Only questions that have definite answers are considered.
    E.g., \texttt{What is the brand of this product?}
(2) \textit{Detailed Description}.
    We ask the assistant to give a comprehensive and detailed desperation of the given product.
    E.g., \texttt{Please describe the product in this image in detail.}
(3)\textit{ Complex Reasoning}.
    The above two types focus on the visual content itself, based on which we further create reasoning questions.
    E.g., \texttt{What occasions is this clothing suitable for?}

\noindent \textbf{Evaluation.}
We follow LLaVA to perform GPT-assisted evaluation. 
After obtaining the responses from models, we feed the question, ground truth text information, and the generated responses, to a judge (GPT-4). The judge evaluates the quality of generated responses from models, and gives an overall score on a scale of 1 to 9.
We report the relative scores w.r.t. the GPT-4 model that has the textural ground truth description as input. Details are available in Appendix~\ref{sec:appendix:D}.
\vspace{-3mm}

\begin{table}[h]

\setlength{\tabcolsep}{4pt} 
\renewcommand{\arraystretch}{0.8} 

\begin{center}
\scalebox{0.65}{%
\centering
\begin{tabular}{lccccc}
\toprule
Model  & Average & Reasoning & Conversation & Detail \\ 
\midrule
LLaVA-v1.5-7B  & 57.9 & 73.2 & 62.8 & 55.4 \\ 
LLaVA + sentence-level RAG  & 59.6 & 74.4 & 64.1 & 57.8 \\ 
\midrule
\rowcolor{gray!15} Ours & \textbf{68.0} & \textbf{78.9} & \textbf{74.4} & \textbf{65.9} \\ 
\bottomrule
\end{tabular}
}
\end{center}
\vspace{-3mm}

\caption{
\small
Results on Fashion-Bench. Sentence-level RAG refers to using retrieved captions as in-context prompts for LLaVA-v1.5-7B.
}
\label{paper_tab_fashion}
\vspace{-3mm}
\end{table}

\noindent \textbf{Results.}
We use a combination of fashion data as our retrieval datastore, including: FashionGen~\cite{rostamzadeh2018fashion} training set, Fashion200k~\cite{han2017automatic} and PolyvoreOutfits~\cite{vasileva2018learning}, resulting in a total of 546.5K image-text pairs. We extract tags of a product from captions.
Results in Tab.~\ref{paper_tab_fashion} demonstrates the effectiveness of TUNA.

\begin{figure}[t]
\centerline{\includegraphics[scale=0.38]{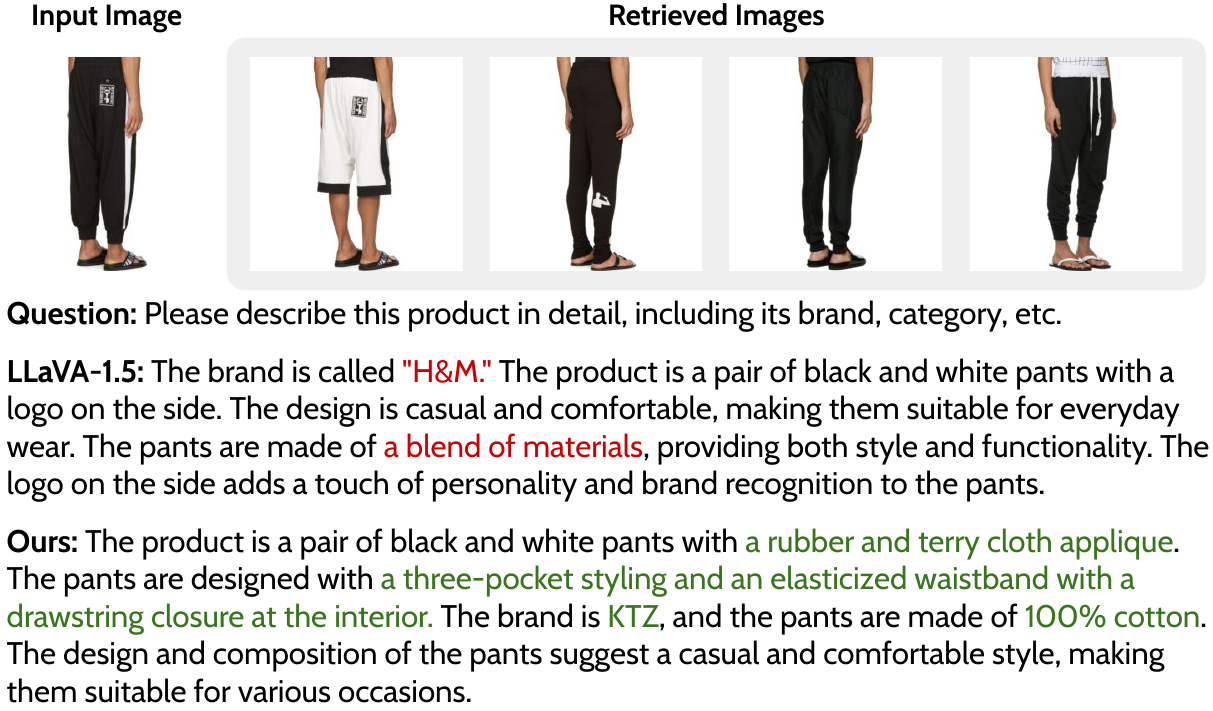}}
\vspace{-2mm}
\caption{
\small
An example on Fashion-Bench. Precise answers in green and vague ones in red.
}
\vspace{-6mm}
\label{fig_paper_fashion}
\end{figure}

\section{Conclusion}
In this paper, we discussed three challenges for MLLMs: (1) mention of non-existent objects, (2) neglect of visual details and (3) failure to identify novel objects and entities, and one of the potential causes: the bottleneck from the image-to-text translation. 
To alleviate these problems, we introduced TUNA, a tag-grounded visual instruction tuning framework with retrieval-augmentation, which achieves competing performance over 12 VQA and multimodal benchmarks, compared to baselines with the same LLM and finetuning data. 

\clearpage
\section*{Limitations}

Being lightweight and effective, our model could be easily further improved with simple modifications to overcome existing limitations. 
Our model is bottlenecked by the capability of CLIP~\cite{radford2021learning}, which can affect our model performance in two ways. First, the quality of retrieved images are highly related to it. As we use tags associated to the retrieved images as additional information, more relevant images we have, more relevant tags we obtain. Second, our adaptive weight tuner also relies on the knowledge of CLIP. 
For instance, even if we obtain a highly relevant tag, e.g., ``Diamond Head'' from the retrieved similar images, if image-text pairs containing ``Diamond Head'' do not exist in the 400M pre-training data of CLIP, CLIP cannot effectively align the text embeddings of ``Diamond Head'' to a photo of diamond head, subsequently, low weights would be assigned to the tag ``Diamond Head'' in our weight tuner, even though it is the ground truth. Fortunately in most cases, CLIP is capable of handling it. If not, we can easily replace CLIP with a more powerful vision-language models.

Our current design of the retriever is also simple, where we retrieve images regardless of the language instruction. A solution could be using Q-former~\cite{li2023blip}, where instruction-aware visual features could be used for retrieval. We leave them for future work.




\bibliography{custom}

\clearpage
\appendix

\section{Extended Related Works}
\label{sec:appendix:A}


\subsection{Retrieval-Augmented Multimodal Learning}
We are distinct from existing works on retrieval-augmented multimodal learning~\cite{ramos2023smallcap, ramos2023retrieval, yang2023re, hu2023reveal, lin2024fine, li2023evcap} in that we are motivated from the object-oriented challenges in visual instruction tuning, which leads to notable differences in (1) target task, (2) motivations, (3) retrieved knowledge and (4) usage of additional information.

Most existing works above focus on image captioning, where short captions (usually one or two sentences) are generated given an input image. While in our case, our model is asked to follow the given instruction, infer from the given image, and often provide a long and detailed response. The difference of tasks therefore lead to different challenges, thus the motivation of using retrieval-augmentation is also distinct. While existing models exploit retrieved captions for general purposes of providing related contents to help the captioning of the current image (e.g., help to better organize the language, or provide additional knowledge on image content or context), in our scenario, the retrieved \textit{tags} aim to provide rich object-aware information to enhance the attention to object details, and help with the object or entity identification. Moreover, the capability of performing tag-grounded generation is enabled during our visual instruction tuning. In addition, we have meticulously crafted novel modules aimed at enriching the representation of retrieved tags and adaptively reallocating the attention to them based on their relevance. 

\subsection{Multimodal Learning with Tags}
We are distinct from existing works~\cite{huang2023tag2text, zhou2020unified,li2020oscar,hu2021vivo,huang2022idea} that introduce object tags as anchor points to help the learning of semantic alignments between images and texts in (1) substantially different objectives, (2) type of used tags and (3) the usage of them.

Existing works~\cite{huang2023tag2text, zhou2020unified,li2020oscar,hu2021vivo,huang2022idea} use tags for the representation learning of semantic alignments between images and texts. 
For instance, OSCAR~\cite{li2020oscar} propose to use object tags to align the object-region features in the pre-trained linguistic semantic space.  
Wu et al.~\cite{wu2016value} utilize solely the predicted object tags as input to an LSTM for image captioning, whereas You et al.~\cite{you2016image} incorporate both tags and region features. In contrast, Zhou et al.~\cite{zhou2020unified} augment region features with the object prediction probability vector, leveraging salient regions identified by object detectors, to enrich the visual input for pre-training. 
In our case, object-oriented tags are used as groundings to provide additional information on the given input image, therefore alleviating neglect of object details and failure to identify novel objects or entities. Besides, the capability of tag-grounded instruction-following in our model is also unique.
The large and abundant annotation-free tags we have (around 3.2M) also makes our work distinctive from the above.
As we want to inform our model of more relevant object-oriented knowledge like object names, object attributes while ignoring less relevant ones, we also design new modules towards this end.

\subsection{Continual Learning of Multimodal Large Language Models}
Continual Learning aims to continuously learn a model from new data in different manners, such as class-incremental~\cite{qi2023better}, data-incremental~\cite{9449913, hua2020less} and domain-incremental~\cite{qi2024extending, zhu2023trustworthy, 10.1145/3657301}.

Zhai et al.~\cite{zhai2023investigating} studies the continual learning of multimodal large language models in the context of object classification. 
They demonstrate that the finetuned popular open-source MLLMs, such as LLaVA~\cite{liu2024visual}, exhibited degraded performance compared to their pre-trained frozen vision encoders, such as CLIP~\cite{radford2021learning}. 
It is an example of the problem caused by the misalignment between the CLIP embeddings of the input image and the LLM text embeddings, as we illustrated in the Introduction Section.

\section{Tag Mining}
\label{sec:appendix:B}

\begin{figure}[h]
\centerline{\includegraphics[scale=0.4]{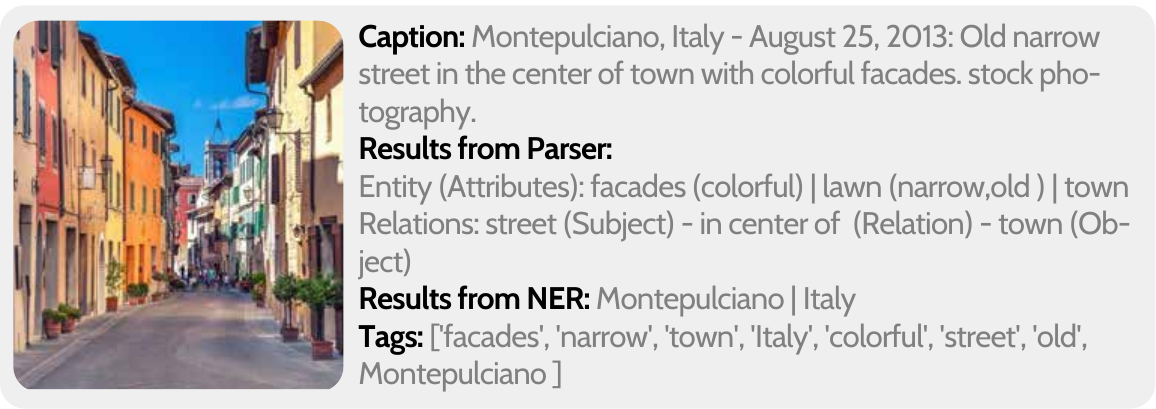}}
\caption{
Examples of tags derived from parsing and NER results.
}
\label{append_fig_parse}
\end{figure}

To mine tags from texts, we parse each caption into a set of tags with a combination of FACTUAL scene graph parser~\cite{li2023factual} and Named Entity Recognition (NER) with spaCy, yielding 3M tags extracted from 15M captions in CC3M~\cite{sharma2018conceptual} and CC12M~\cite{changpinyo2021conceptual}. We show several examples in Fig.~\ref{append_fig_parse}. 

Given that the FACTUAL scene graph parser~\cite{li2023factual} is built on a large language model, there is a slight probability that it may produce nonsensical lengthy sequences. We employ a filtering mechanism to exclude tags exceeding 30 characters in length. 

\section{Training Details}
\label{sec:appendix:C}
\subsection{Datasets}
LLaVA-665K~\cite{liu2023improved} is collected and built with a variety of datasets, containing VQA, OCR, region-level VQA, visual conversation and language conversation data. 
In ShareGPT4V~\cite{chen2023sharegpt4v}, the supervised fine-tuning captions were collected from GPT4-Vision. 
Following Chen et al.~\cite{chen2023sharegpt4v}, a corresponding portion of detailed captions in the Supervised Fine-Tuning (SFT) datasets (i.e., LLaVA-665K) is replaced with a selection from the 100K GPT4-Vision-generated captions.

\subsection{Hyperparameter}
We follow the hyperparameter setting in LLaVA-1.5~\cite{you2016image}. Details are summerized in Tab.~\ref{tab:hyperparameters}.

\begin{table}[t]
\begin{center}
\scalebox{0.8}{%
\setlength{\tabcolsep}{20pt}
\centering
\begin{tabular}{lc}
\toprule
\multicolumn{2}{c}{Hyperparameters} \\ 
\midrule
Batch Size & 128  \\ 
Learning Rate & $2 \times 10^{-5}$  \\ 
Learning Rate Schedule & Cosine Decay  \\ 
Learning Rate Warmup Ratio & 0.03  \\ 
Weight Decay & 0  \\ 
Epoch & 1  \\ 
Optimizer & AdamW  \\ 
DeepSpeed Stage & 3  \\ 
\bottomrule
\end{tabular}
}
\end{center}
\caption{Hyperparameters for Instruction Finetuning.}
\label{tab:hyperparameters}
\end{table}

\section{Zero-Shot Inference on Fashion Data}
\label{sec:appendix:D}

\subsection{Fashion-Bench}

To explore the effectiveness of TUNA on OOD data from another specific domain, we further collect data from FashionGen~\cite{rostamzadeh2018fashion} validation set and create a benchmark to measure the model’s instruction-following capability in Fashion domain. 
Following LLaVA~\cite{liu2024visual}, we leverage GPT-4 to measure the quality of generated responses.
Specifically, we create triplets consisting of image, ground-truth textual descriptions, and question.
The candidate models (e.g., TUNA, LLaVA) predict the answers based on the question and the image.
To provide an approximate upper bound, we build a reference prediction based on the question and
the ground-truth textual descriptions, using the text-only GPT-4, following Liu et al.~\cite{liu2024visual}
.
After obtaining the responses from both models, we feed the question, visual information (in the format of textual descriptions), and the generated responses from both assistants, to the judge (i.e., text-only GPT-4). The text-only GPT-4 evaluates the helpfulness, relevance, accuracy, and level of detail of the responses from the assistants, and gives an overall score on a scale of 1 to 9, where a higher score indicates better overall performance.
We report relative scores w.r.t. the text-only GPT-4 model that uses the textural ground truth
description as visual input. 

Similar to LLaVA-Bench (In-the-Wild)~\cite{liu2024visual}, we also collect a set of 24 images from FashionGen~\cite{rostamzadeh2018fashion} validation set, with one question associated with each image.
The questions are from one of the three types:
\begin{enumerate}
    \item Conversation.
    We design a conversation between the assistant and a person asking questions about the product. A diverse set of questions are asked about the content of the image, including the product brands, categories, materials, etc. Only questions that have definite answers are considered.
    E.g., \texttt{What is the brand of this product?}
    \item Detailed Description. 
    We ask the assistant to give a comprehensive and detailed desperation of the given product.
    E.g., \texttt{Please describe the product in this image in detail.}
    \item Complex Reasoning. 
    The above two types focus on the visual content itself, based on which we further create reasoning questions.
    E.g., \texttt{What occasions is this clothing suitable for?}
\end{enumerate}

\begin{figure*}[h]
\centerline{\includegraphics[scale=0.55]{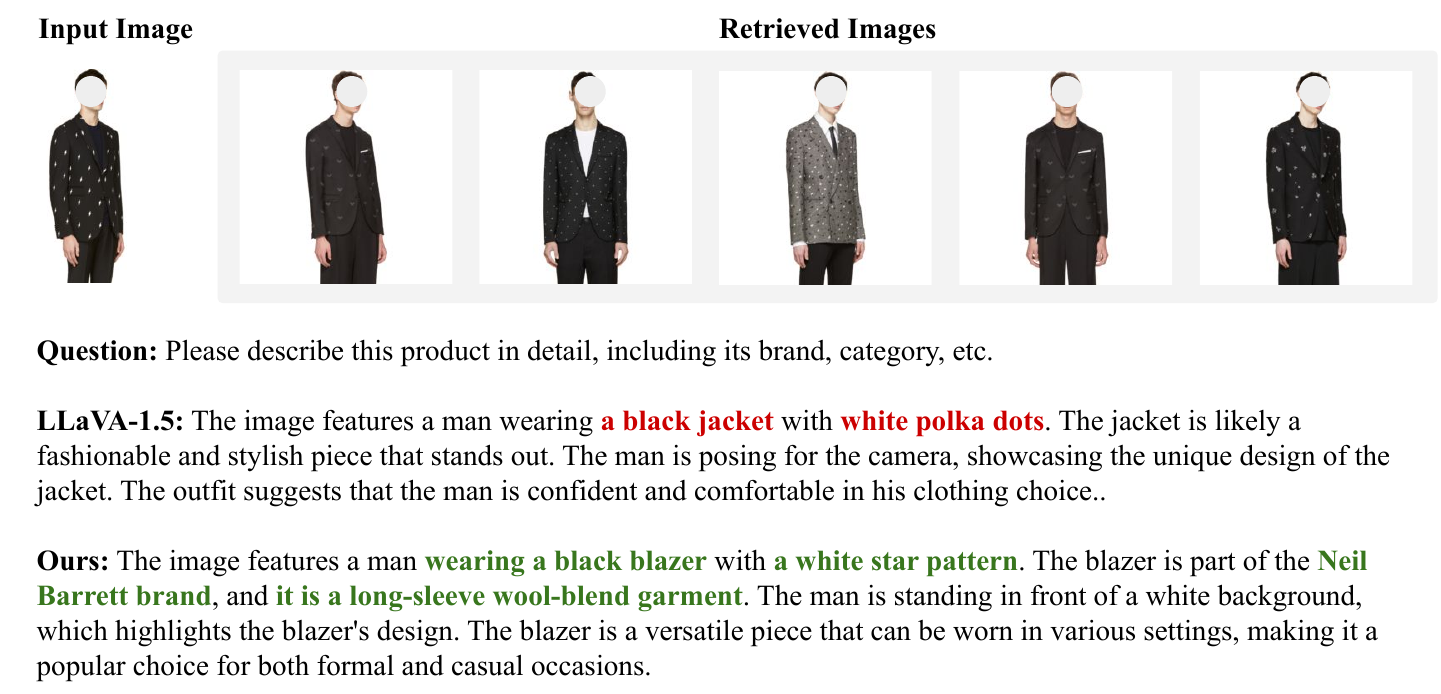}}
\caption{
\small
Examples of results on Fashion-Bench. Precise and accurate descriptions are marked green while vague or wrong descriptions are marked red.
}
\label{fig_fashion_introduce}
\end{figure*}


\subsection{Experiments}
\begin{table}[h]

\setlength{\tabcolsep}{2pt} 
\renewcommand{\arraystretch}{0.8} 

\begin{center}
\scalebox{0.8}{%
\centering
\begin{tabular}{lccccc}
\toprule
Model  & Average & Reasoning & Conversation & Detail \\ 
\midrule
LLaVA-v1.5-7B  & 57.9 & 73.2 & 62.8 & 55.4 \\ 
\rowcolor{gray!15} Ours & \textbf{68.0} & \textbf{78.9} & \textbf{74.4} & \textbf{65.9} \\ 
\bottomrule
\end{tabular}
}
\end{center}
\caption{Results on Fashion-Bench. Our model consistently outperforms the baseline.}
\label{tab_fashion}
\end{table}

We use a combination of fashion data as our retrieval datastore, including: FashionGen~\cite{rostamzadeh2018fashion} training set, Fashion200k~\cite{han2017automatic} and PolyvoreOutfits~\cite{vasileva2018learning}, resulting in a total of 546.5K image-text pairs.
To obtain the tags of a product, we extract them from the caption or associated product specifications (e.g., brand) of the product.

Results in Tab.~\ref{tab_fashion} demonstrates the effectiveness of TUNA, especially on `Conversation' and `Detail', where retrieved tags on product specifications are very helpful to identify the related details of the input product. Examples are available in Fig.~\ref{fig_fashion_introduce} and Fig.~\ref{fig_fashion}.

\section{More Examples}
\label{sec:appendix:E}
\begin{figure*}[t]
\centerline{\includegraphics[scale=0.59]{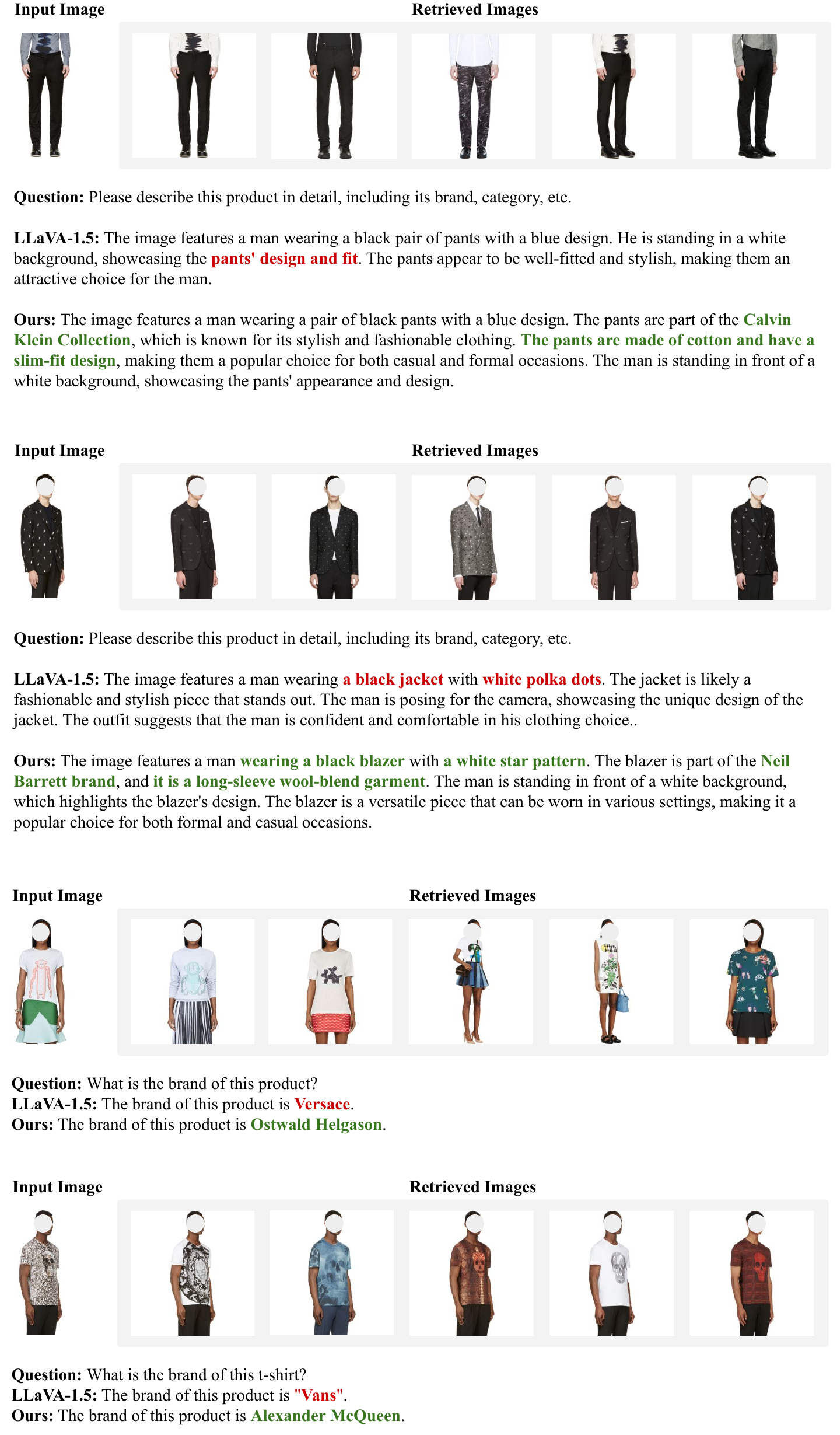}}
\caption{
\small
Examples of results on Fashion-Bench. Precise and accurate descriptions are marked green while vague or wrong descriptions are marked red.
}
\label{fig_fashion}
\end{figure*}

\begin{figure*}[t]
\vspace{-1mm}
\centerline{\includegraphics[scale=0.55]{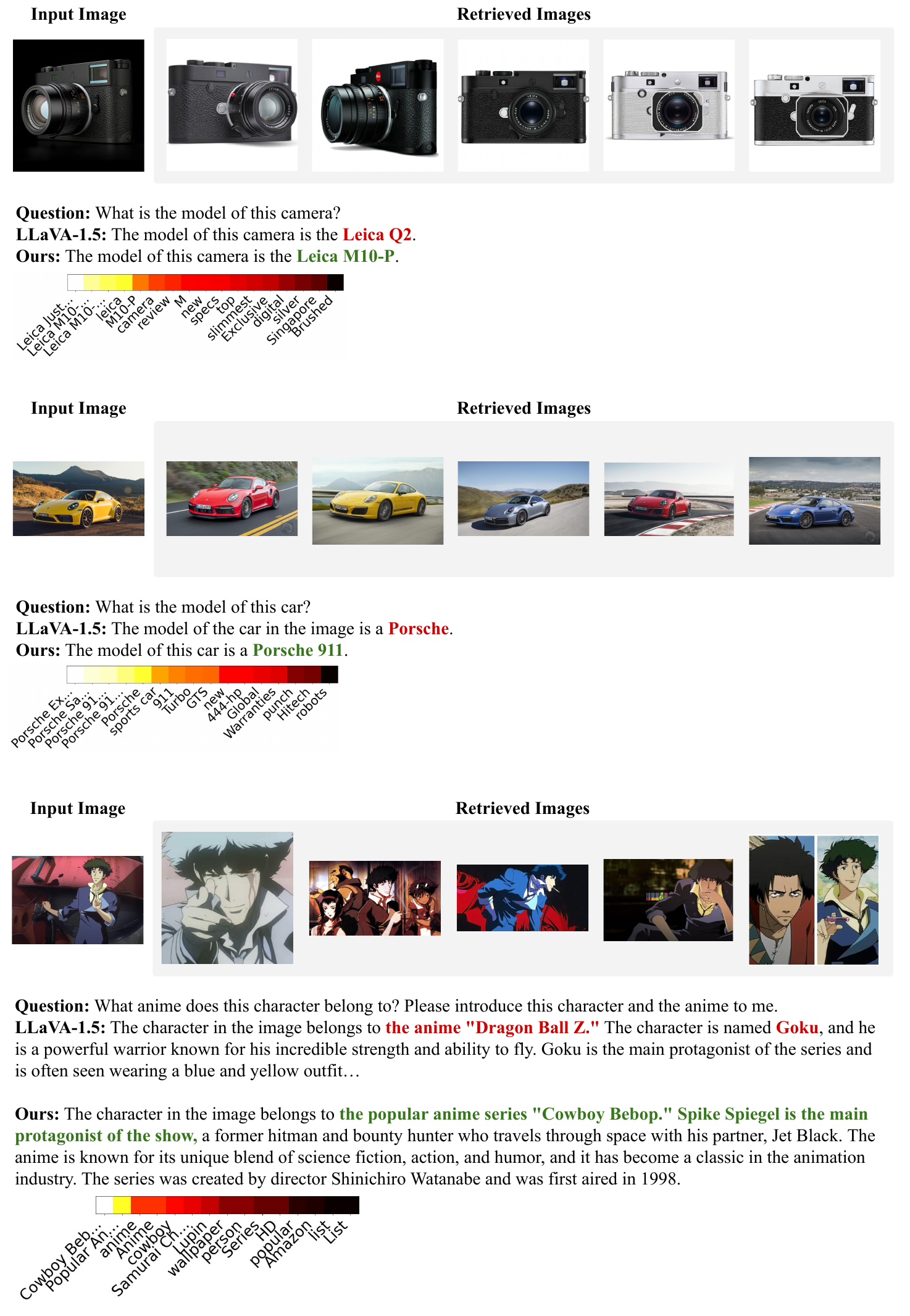}}
\caption{
\textbf{TUNA examples.} For each example, we show top 5 retrieved images. We show the entire tag set associated with all retrieved images as well as their tuned weights in heatmap, where the brightest region for the highest weight 1 and darkest region for the lowest weight 0 (Zoom in for better view). Correct and precise answers are marked green while vague or wrong ones in red.
}
\label{fig_tuna_example}
\end{figure*}

We present more examples with TUNA and LLaVA-1.5 in Fig.~\ref{fig_tuna_example} and Fig.~\ref{fig_fashion}.
In Fig.~\ref{fig_tuna_example}, we provide Out-of-Distribution (OOD) images of real-world products or television works, and ask TUNA and LLaVA-1.5 to provide answers to the question. 
In Fig.~\ref{fig_fashion}, we provide  Out-of-Distribution (OOD) images in fashion domain, and ask the models to provide answers to the question. 

When provided with OOD images, where novel objects or entities often appear, LLaVA-1.5 fails to correctly or precisely identify them due to a limited number of training samples. Although the CLIP vision encoder, which is pre-trained with over 400M samples, can effectively extract their visual features, the multimodal connector cannot effectively map them to text embeddings input to the LLM.
In contrast, TUNA is effective in identifying unseen objects or entities, as the input OOD image is directly mapped to a set of retrieved tags from a large-scale external datastore, which has a better coverage of OOD data. 

In examples in Fig.~\ref{fig_fashion}, where specific in-domain knowledge, i.e., fashion domain, is required for give a detailed and precise description of the given product, such as its brand, design, or composition (material), LLaVA fails to correctly identify them or response with detailed descriptions on them. 

For instance, in the example in Fig~\ref{fig_fashion_introduce}, the only useful information about the given product itself is ``a black jacket with white polka dots'', where LLaVA-1.5 fails to precisely describe it as a ``blazer''. Moreover, LLaVA-1.5 does not mention its design and brand even if we explicitly ask it the brand of this product.
In contrast, TUNA precisely describes its design details, style and the brand, benefiting from the retrieved products which are similar to the input product in design, brand, category or style. TUNA could effectively refer to the retrieved tags and learn from the useful ones with our tag encoder.

Cases are similar in examples from Fig~\ref{fig_tuna_example}, where TUNA correctly identifies the novel object in the input image with retrieved knowledge. Meanwhile, LLaVA-1.5 fails to identity the model of the Leica camera, Porsche car, and the name of the character and anime in the input images.

\section{Full Experiment Results}
\label{sec:appendix:F}

We show the full results on POPE in Tab.~\ref{appen_tab_POPE}.

\begin{table*}[t]

\setlength{\tabcolsep}{8pt} 
\renewcommand{\arraystretch}{0.8} 

\begin{center}
\scalebox{0.75}{%

\begin{tabular}{l|l|cccccccc}
\toprule
Datasets & Metrics & \cellcolor{gray!15}Ours & Ferret & Shikra & InstructBLIP & MiniGPT4 & LLaVA & MM-GPT & mPLUG-Owl \\
\midrule 
\multirow{4}{*}{ Random } & Accuracy $(\uparrow)$ & \cellcolor{gray!15}\textbf{91.00} & 90.24 & 86.90 & 88.57 & 79.67 &88.00 & 50.10 & 53.97 \\
& Precision $(\uparrow)$ & \cellcolor{gray!15}\textbf{98.05} &97.72 & 94.40 & 84.09 & 78.24 & 97.44 & 50.05 & 52.07 \\
& Recall $(\uparrow)$ &\cellcolor{gray!15}84.10 & 83.00 & 79.26 & 95.13 & 82.20 & 78.80 & \textbf{100.00} & 99.60 \\
& F1 Score $(\uparrow)$ &\cellcolor{gray!15}\textbf{90.93} & 89.76 & 86.19 & 89.27 & 80.17 & 87.13 & 66.71 & 68.39 \\
\midrule 
\multirow{4}{*}{ Popular } & Accuracy $(\uparrow)$ & \cellcolor{gray!15}\textbf{90.16} & 84.90 & 83.97 & 82.77 & 69.73 & 87.43 & 50.00 & 50.90 \\
& Precision $(\uparrow)$ & \cellcolor{gray!15}\textbf{95.46} & 88.24 & 87.55 & 76.27 & 65.86 & 95.24 & 50.00 & 50.46 \\
& Recall $(\uparrow)$ &\cellcolor{gray!15}84.20 & 80.53 & 79.20 & 95.13 & 81.93 &78.80 & \textbf{100.00} & 99.40 \\
& F1 Score $(\uparrow)$ & \cellcolor{gray!15}\textbf{90.56 }& 84.21 & 83.16 & 84.66 & 73.02 & 86.24 & 66.67 & 66.94 \\
\midrule 
\multirow{4}{*}{ Adversarial } & Accuracy $(\uparrow)$ & \cellcolor{gray!15}\textbf{88.43} & 82.36 & 83.10 & 72.10 & 65.17 & 85.50 & 50.00 & 50.67 \\
& Precision $(\uparrow)$ & \cellcolor{gray!15}\textbf{91.99} & 83.60 & 85.60 & 65.13 & 61.19 & 90.99 & 50.00 & 50.34 \\
& Recall $(\uparrow)$ & \cellcolor{gray!15}84.20 & 80.53 & 79.60 & 95.13 & 82.93 & 78.80 & \textbf{100.00} & 99.33 \\
& F1 Score $(\uparrow)$ & \cellcolor{gray!15}\textbf{87.63} & 82.00 & 82.49 & 77.32 & 70.42 & 84.45 & 66.67 & 66.82 \\
\midrule 
\multicolumn{2}{c|}{Average F1} & \cellcolor{gray!15}\textbf{89.50} & 85.32 & 83.94 &  83.75 & 74.53 & 85.94 & 66.68 &  67.38 \\
\bottomrule
\end{tabular}
}
\end{center}
\caption{ Results on POPE. We outperform competing baselines including Ferret~\cite{you2023ferret}, which is finetuned on grounding and referring data. }
\vspace{-4mm}
\label{appen_tab_POPE}
\end{table*}

\section{Benchmarks}
\label{sec:appendix:G}
We compare TUNA with SoTA methods on 12 benchmarks, including five VQA benchmarks: $\text{VQA}^{\text{v2}}$~\cite{goyal2017making}, GQA~\cite{hudson2019gqa}, VizWiz~\cite{gurari2018vizwiz}, ScienceQA-Image ($\text{SQA}^{\text{I}}$)~\cite{lu2022learn}, TextVQA ($\text{VQA}^{\text{T}}$)~\cite{singh2019towards}, and seven more recently multimodal benchmarks designed for LLMs: POPE~\cite{li2023evaluating}, 
MME (Fu et al., 2023), 
MMBench (MMB)~\cite{liu2023mmbench}, MMBench-Chinese ($\text{MMB}^{\text{CN}}$)~\cite{liu2023mmbench}, SEED~\cite{li2023seed}, LLaVA-in-the-Wild ($\text{LLaVA}^{\text{W}}$)~\cite{liu2023improved}, and MM-Vet~\cite{yu2023mm}.

$\text{VQA}^{\text{v2}}$~\cite{goyal2017making} and VizWiz~\cite{gurari2018vizwiz} are benchmarks for traditional Visual Question Answering (VQA) tasks.
MME~\cite{fu2023mme} evaluates LLMs’ assesses and cognition capabilities through a wide range of carefully crafted questions across 14 sub-tasks. 
MMBench (MMB) and MMBench-Chinese ($\text{MMB}^{\text{CN}}$)~\cite{liu2023mmbench} benchmarks manually design questions to evaluate the LLM’s visual reasoning and perception abilities in English and Chinese, respectively. 
SEED~\cite{li2023seed} generated a dataset comprising around 19K questions with images and videos with the GPT4 assistance.

\section{Analysis on Choices of Datastores}
\label{sec:appendix:H}
From Tab.~\ref{tab_aba} and previous analysis we know that the quality of retrieved tags is critical. Therefore, the datastore, where the images are (with corresponding tags) retrieved from is crucial. Here we study how different choices of datastores can affect the model performance. 
In the default setting, we use a combination of CC12M~\cite{changpinyo2021conceptual}, CC3M~\cite{sharma2018conceptual} and COCO training set~\cite{lin2014microsoft}. 
Two of the three retrieval datasets, CC3M and the COCO training set, share overlaps with the LLaVA training data, which is a frequent scenario in retrieval-augmented generation, where a datastore with full or partial overlap with the training data is common~\cite{ramos2023smallcap, ramos2023retrieval, yang2023re, hu2023reveal, lin2024fine, li2023evcap}.
While CC12M and CC3M are different in size but similar in content style, COCO is different from them in both size and content. CC12M and CC3M consist of web image-text pairs, where the variance in caption quality and style is more significant.
In COCO, captions are human-written, where the language style is more coherent, usually a short and plain description of the image. Consequently, tags extracted from COCO captions are often commonly used words and phrases and are very general, for instance, ``boy'', ``girl'', ``plane'' and ``train'', etc. It can provide the existence of objects in the image, which might help to alleviate the mention of non-existent objects. However, it is hard help to improve object or entity identification as these commonly seen phrases are very likely to be already included in LLaVA training data and new retrieval mappings cannot be established.
On the contrary, CC12M and CC3M provide an ocean of novel objects and entities, which could greatly improve the image-to-text translation process with additional new retrieval mappings built from them.

We are curious to see how different datastore size and datastore style can influence our model performance. In additional to the default setting, we perform the tag-grounded instruction tuning with different datastores, and use them for retrieval during inference, respectively. Results are available in Tab.~\ref{tab_aba_datastore}.

It is not surprising that the default setting with largest datastore size consistently outperforms other baselines.
In most cases, the baseline with CC12M is the second best one while the one with COCO training set performs worst,
except for on POPE. This is because POPE is built with COCO validation set, which shares the same style of the COCO training set.
On other multimodal benchmarks, the improvements with COCO training set is less than CC12M and CC3M.
Particularly, in LLaVA-in-the-Wild (LLaVA-W) benchmark, where all test images are not overlapped with COCO training and validation set, COCO training set as datastore does not help at all.

\end{document}